# Diagnosing Representation Dynamics in NER Model Extension

Xirui Zhang, Philippe de La Chevasnerie, Benoit Fabre, papernest (www.papernest.com, 157 boulevard Macdonald 75019 Paris)

1. Introduction

Being able to accurately identify and protect Personally Identifiable Information (PII) is critical for AI systems processing human language. This challenge is particularly pronounced for text from Automatic Speech Recognition (ASR) systems, which is inherently noisy with informalities, grammatical errors, and non-standard entity representations. While pre-trained language models like BERT perform well on canonical Named Entity Recognition (NER) benchmarks, reliable PII detection in noisy, real-world contexts requires specialized fine-tuning.

A common operational requirement is expanding NER systems to recognize new entity types. For instance, a model trained on standard entities—persons (PER), locations (LOC), and organizations (ORG)—may later need to detect domain-specific PII like email addresses or phone numbers. Conventional wisdom in continual learning suggests that extending models with new classes degrades performance on previously learned categories due to catastrophic forgetting or representational interference. This phenomenon occurs as the model adjusts its feature space to accommodate new concepts, potentially disrupting the decision boundaries of old ones.

Our initial experiments revealed a surprising result: when jointly fine-tuning a pre-trained BERT model on both standard entities and new spoken-format PII entities, performance on original PER, LOC, and ORG tags remained stable with minimal degradation. This peaceful coexistence —a counter-intuitive finding— motivates our central research question: Why does adding new PII entities not significantly impact existing entity recognition?

We investigate the underlying reasons for this phenomenon.We hypothesize that stability stems from fundamental differences between entity types: standard entities like PER and ORG are learned through semantic and contextual cues, while PII entities are identified through lexical and morphological patterns. These feature types may occupy largely independent learning mechanisms within the model's representation, minimizing interference.

To probe this hypothesis and explore the model's behavior, we employ an incremental learning setup not as a performance goal, but as a diagnostic tool. By simulating the sequential introduction of new entities to a pre-trained model, we can amplify the subtle points of friction and competition between entity representations. This diagnostic approach leads us to two key discoveries:

Representation Overlap as a Vulnerability: Under the pressure of incremental learning, the LOC entity consistently emerges as the most vulnerable among the original classes. We demonstrate that this is likely due to a feature overlap, as locations (e.g., addresses with numbers, postal codes) may share surface-form characteristics with pattern-based PII.

Reverse O-tag representation drift: The challenge of O-tag representation drift is a known issue in NER incremental learning. This typically occurs when new training data contains labels only for new entity types, while instances of old entities are labeled as 'O', which confuses the model. In our work, we observe the inverse. Our initial training phase includes instances of the new entities but masks them as 'O'. When the model is later trained on a small dataset rich in these new entities, it faces profound confusion between the newly introduced labels and the ingrained 'O' classification. Our diagnostic experiments reveal that simply freezing all old classifier heads renders the model completely incapable of learning the new entities. However, by selectively unfreezing the 'O' tag's classifier, the model successfully begins to learn. This reverse drift requires the background class representation to adapt by releasing previously absorbed entity patterns, fundamentally altering the learned distinction between entities and non-entities.

Our findings provide diagnostic insights into how NER models handle capability expansion, revealing the underlying mechanisms that enable or hinder learning of new entity types. This article contributes not to a new state-of-the-art incremental learning algorithm, but rather a scientific inquiry into the internal dynamics of NER models when faced with evolving requirements.

## 2. Related Works

### 2.1 Class-Incremental Learning

Class-Incremental Learning (CIL) addresses the challenge of sequentially learning new classes without forgetting previously acquired knowledge, a phenomenon known as catastrophic forgetting (McCloskey and Cohen, 1989). A central goal in CIL is to balance the stability-plasticity dilemma, enabling a model to both retain old information and acquire new skills. Mainstream approaches can be broadly categorized into several families.

Replay-based methods, also known as rehearsal (Robins, 1995), have emerged as one of the most effective strategies for mitigating catastrophic forgetting. This approach can be broadly divided into two main categories. The first, exemplar replay, involves storing a small subset of representative samples from previous training, known as exemplars. These exemplars are then mixed with new data during training to revisit previous knowledge. Seminal works like iCaRL (Rebuffi et al., 2017) and LUCIR (Hou et al., 2019) have demonstrated the power of combining this technique with knowledge distillation to maintain performance. Methods such as GEM (Lopez-Paz & Ranzato, 2017) and A-GEM (Chaudhry et al., 2018) leverage exemplars to constrain the gradient updates, ensuring that learning new tasks does not increase the loss on previous ones. The second category, synthetic replay, addresses the practical concerns of privacy, storage, and data access associated with storing real data. Instead of keeping original samples, this approach trains a generative model to synthesize data from past distributions. A pioneering work in this area by Shin et al. (2017) utilized a generative model to create pseudo-samples for rehearsal.

Regularization-based methods introduce a penalty term to the loss function to prevent significant changes to parameters crucial for previous tasks. A foundational approach in this category is Elastic Weight Consolidation (EWC), which uses the Fisher information matrix to estimate the importance of each parameter for tasks learned previously (Kirkpatrick et al., 2017). However, a key drawback of EWC is its assumption that the Fisher matrix is diagonal, a condition that is rarely true. Addressing this, Liu et al. (2018) proposed a method to rotate the network's parameter space to better satisfy this diagonal assumption, thereby making weight consolidation more effective. Researchers have also developed alternative metrics for quantifying parameter importance. For example, Synaptic Intelligence (SI) computes importance in an online manner by measuring the contribution of each parameter to the change in the loss function throughout the entire training trajectory (Zenke et al., 2017). Taking a different approach, Memory Aware Synapses (MAS) calculates importance by measuring how sensitive the predicted output function is to a change in a parameter, which uniquely allows importance to be updated using unlabeled data (Aljundi et al., 2018). Another prominent technique is knowledge distillation (KD), pioneered in CIL by Learning without Forgetting (LwF) (Li and Hoiem, 2018), which forces the student model to mimic the output logits of the teacher model (the model from the previous step).

Architecture-based methods directly address catastrophic forgetting by modifying the network's structure to accommodate new tasks, thereby physically isolating knowledge representations. This paradigm is primarily divided into two main strategies. The first, parameter isolation, operates within a fixed-capacity network by dedicating distinct subsets of parameters to different tasks. Foundational works like PackNet (Mallya and Lazebnik, 2018) achieve this by iteratively pruning weights to "pack" knowledge, while Hard Attention to the Task (HAT) (Serrà et al., 2018) learns binary masks to freeze task-specific pathways. The second strategy involves dynamic network expansion, where the model's capacity grows with each new task. Progressive Neural Networks (PNN) (Rusu et al., 2016) pioneered this approach by adding a new network column for each task, effectively preventing forgetting at the cost of linear model growth. More recent methods like DER (Yan et al., 2021) and FOSTER (Wang et al., 2022a) refine this by dynamically expanding the feature extractor and then employing knowledge consolidation to manage model size. The rise of Pre-trained Models (PTMs) has revitalized this area through parameter-efficient techniques, such as adding lightweight adapters or prompts (Wang et al., 2022d), which allows for knowledge expansion while keeping the large backbone model frozen.

## 2.2 Incremental Learning for Named Entity Recognition (INER)

While drawing heavily from general CIL, INER presents unique challenges, primarily due to the nature of sequence labeling. A key issue is the "unlabeled entity problem", where tokens belonging to old classes are labeled as "O" (non-entity) in the training data for new classes. This contaminates the "O" tag's semantics, leading to confusion between true non-entities and previously learned entities (Ma et al., 2023b; Wu et al., 2025a).

Early and influential work by Monaikul et al. (2021) systematically adapted the knowledge distillation framework for INER, using the old model as a teacher to provide soft labels for unlabeled tokens in new data streams. To address the scarcity of old-class entities in new data, subsequent works have explored generating synthetic data. For instance, Wang et al. (2022b) proposed a data-free approach by inverting the old NER model to reconstruct synthetic samples for distillation, while Xia et al. (2022) introduced a "Learn-and-Review" framework that uses a generator to create synthetic contexts for review. Most recently, a paradigm shift has emerged, reformulating INER as a unified seq2seq generation task. This approach, exemplified by Wu et al. (2025b), leverages dynamic prefix-tuning to guide a frozen language model, enabling efficient knowledge acquisition for new entity types while preserving old knowledge through stable prefixes.

## 2.3 Mechanistic Analysis of Incremental Learning

Beyond developing new algorithms, a parallel line of research, though less frequent, seeks to understand the underlying causes and mechanisms of catastrophic forgetting.

Bell and Lawrence (2022) employed behavioral research methods from experimental psychology, using carefully designed synthetic tasks to manipulate variables such as perceptual similarity and semantic similarity between sequential tasks. Through controlled experiments, they demonstrated that catastrophic forgetting is most severe when tasks are perceptually similar but semantically conflicting.

To understand the mechanisms underlying forgetting, researchers have worked to quantify and localize the parameter changes responsible for performance degradation. Wiewel and Yang (2019) proposed a method to attribute the contribution of individual parameters to CF, finding that layers closer to the output are more susceptible to destructive updates. From a representation perspective, Wu et al. (2025a) analyzed the phenomenon of semantic drift within the feature space, showing that forgetting manifests as shifts in the mean and covariance of old class feature distributions within the embedding space. This frames catastrophic forgetting as a problem of maintaining distributional stability. Bell and Lawrence (2022) also identified a correlation between the magnitude of initial gradient updates on new tasks and subsequent forgetting, indicating that early updates are predictive of long-term retention. Lancheros et al. (2023) compared two training regimes: direct fine-tuning on concatenated data versus continual learning. Although their primary goal was to develop a cross-lingual NER system, their comparison also revealed fundamental differences in incremental learning behavior.

These behavioral mechanisms are particularly evident in challenging settings such as Few-Shot Class-Incremental Learning (FSCIL). Zhao et al. (2023) analyzed the hubness phenomenon in FSCIL, where base class prototypes become geometric hubs in feature space and disproportionately attract samples from new classes. Their findings highlight the importance of jointly considering feature representations and classification metrics when analyzing forgetting.

Our research is positioned within the mechanistic analysis of incremental learning. Rather than proposing a new state-of-the-art algorithm, our work contributes to the line of inquiry that prioritizes understanding model behavior, aligning with the diagnostic framework of Bell and Lawrence (2022) and Wiewel and Yang (2019). We employ an incremental learning setup as a framework for a diagnostic ablation study. By systematically altering training configurations we can isolate and analyze the internal mechanisms that enable or inhibit the acquisition of new entity types.

To quantify internal model changes, we adopt the analytical framework of semantic drift, inspired by the work of Wu et al. (2025a). While Wu et al. (2025a) use their analysis of mean and covariance shifts to design novel loss functions for training, our work utilizes this framework as a post-hoc diagnostic tool. We apply these metrics after training to measure the representational consequences of each experimental condition. This approach allows us to probe the model's internal dynamics and identify factors affecting representational stability and interference.

## 3. Experiments and results

### 3.1. Dataset and Entity Types

Our study is conducted on a private dataset of transcribed spoken language in French. The data is characterized by conversational language and contains annotations for both standard and PII-related entities: Person (PER), Location (LOC), Organization (ORG), Phone Number (PHONE), Email Address (EMAIL), IBAN, and domain-specific ID (PDL).

For our experiments, the data is structured into two distinct sets to simulate a real-world model extension scenario:

- Dataset-A-Original: Primary training corpus containing all entity types. For the initial stage of our incremental learning simulation, we use a version where only the standard PER, LOC, and ORG tags are used as ground truth; other PII entities are masked and labeled as the background 'O' tag.
- Dataset-B-Additional: Smaller corpus rich in new PII entities (PHONE, EMAIL, IBAN, PDL). Instances of original entities, though rare, are correctly labeled rather than as 'O'.

### 3.2. Model and Experimental Configurations

Our experiments are based on the Jean-Baptiste/camembert-ner model, a French RoBERTa-based model pre-trained for NER. We evaluate several training configurations.

- Training on BERT Tags: Baseline model trained on Dataset-A-Original using only PER, LOC, ORG labels, treating PII entities as 'O'.
- Joint Training (Upper Bound): Model trained on combined Dataset-A-Original and Dataset-B-Additional with ground-truth labels for all seven entity types from start.
- Naive Incremental Learning: Direct update to the baseline model on Dataset-B-Additional, designed to expose catastrophic forgetting and interference.
- Incremental with O-Tag Adaptation: Identical to Naive Incremental Learning but freezing PER, LOC, ORG classifier heads during second training stage while keeping new PII entities and 'O' tag trainable.

In addition to these main experiments, we conducted two further diagnostic probes to understand the learning constraints. We found that in scenarios where we (a) froze the entire CamemBERT model and only trained the new entity classifiers, or (b) froze all old classifier heads including 'O' while fine-tuning the base model and the new entity classifiers, the model completely failed to learn the new entity types. These results highlight the critical roles of both representation plasticity and background class adaptability, which we will analyze in the following section. Together, the Naive Incremental and the Incremental with O-Tag Adaptation experiments, along with our probes into freezing different model components, form a comprehensive ablation study. This study is not aimed at finding the best performance, but at systematically isolating the effects of representation plasticity and classifier head adaptability to diagnose the causes of learning failure.

| Experiment | F1 overall | F1 LOC | F1 PER | F1 ORG | F1 PHONE | F1 EMAIL | F1 IBAN | F1 PDL |
|---|---|---|---|---|---|---|---|---|
| Training on BERT tags | | 63.11% | 62.03% | 83.63% | — | — | — | — |
| Joint training | 67.76% | 62.20% | 63.76% | 83.78% | 61.54% | 33.3% | 41.3% | 32.56% |
| Naive incremental | 54.96% | 42.24% | 53.85% | 79.03% | 59.86% | 28.39% | 21.25% | 7.8% |
| Incremental + freeze old classifier heads except O | 59.63% | 60.32% | 61.14% | 81.99% | 58.67% | 30.12% | 33.30% | 8.45% |

Table 1 Micro F1 scores of all models

## 3.3. Quantifying Semantic Drift

In addition to the ablation studies that isolate the effects of different training strategies, we employ a quantitative analysis of the model's feature space to diagnose the internal mechanisms of learning and forgetting. We adapt the semantic drift framework from Wu et al. (2025a), who originally utilized these statistical moments to construct novel loss functions for calibrating representational shifts during training. Our work leverages this framework purely as a diagnostic tool. By applying it post-hoc, we can measure the consequences of each training configuration on the model's internal representations without altering the training dynamics themselves. We utilize the following three metrics:

- Mean Drift: Following the methodology of Wu et al. (2025a), we measure the shift in a class's feature centroid by calculating the L2 norm of the weighted average change in token embeddings. This metric quantifies the magnitude of displacement of a class's core representation.
- Covariance Drift: This metric, adapted from Wu et al. (2025a), is used to understand changes in the shape of a class's feature distribution. It is calculated as the mean absolute difference in Mahalanobis distances between pairs of embeddings before and after a training step. A high value signifies that the learned feature combinations defining a class have been altered, indicating representational instability.
- Variance Change: As a complementary metric, we introduce the change in the trace of the covariance matrix. This provides a measure of the feature distribution's volume. A positive value (expansion) suggests that representations have become more dispersed, while a negative value (contraction) signifies that representations have become more compact.

By analyzing these three metrics, we can construct a detailed narrative of the model's internal dynamics, distinguishing between different modes of representational change for each entity class.

Table 2 Comparison of mean drift, covariance drift and variance change

Original vs. joint

| class | Mean drift | Var change | Cov drift |
| LOC | 2.1555 | - 0.2538 | 0.2229 |
| ORG | 2.1977 | - 4.7861 | 0.2869 |
| PER | 3.5152 | - 1.4799 | 0.2238 |
| O | 2.2972 | - 0.0206 | 0.1519 |

Original vs. naive

| class | Mean drift | Var change | Cov drift |
|---|---|---|---|
| LOC | 2.8185 | 4.5992 | 1.0054 |
| ORG | 1.2799 | - 1.0340 | 0.2948 |
| PER | 2.9626 | 6.8460 | 0.4609 |
| O | 1.2826 | - 1.2003 | 1.0886 |

Original vs. freeze except O

| class | Mean drift | Var change | Cov drift |
|---|---|---|---|
| LOC | 2.2926 | - 1.6716 | 0.2157 |
| ORG | 0.9502 | - 1.6130 | 0.2060 |
| PER | 2.0098 | 3.0644 | 0.3588 |
| O | 1.2913 | - 1.2992 | 1.0838 |

Joint vs. naive

| class | Mean drift | Var change | Cov drift |
|---|---|---|---|
| LOC | 4.0083 | 4.8530 | 1.2111 |
| ORG | 2.5168 | 3.7521 | 0.7361 |
| PER | 4.8173 | 8.3260 | 0.8863 |
| O | 2.3471 | - 1.1798 | 0.9130 |

Joint vs. freeze except o

| class | Mean drift | Var change | Cov drift |
| --- | --- | --- | --- |
| LOC | 2.2144 | - 1.4177 | 0.2869 |
| ORG | 2.2324 | 3.1731 | 0.5927 |
| PER | 3.1349 | 4.5442 | 0.7876 |
| O | 2.3580 | - 1.2786 | 0.9045 |

Naive vs. freeze except O

| class | Mean drift | Var change | Cov drift |
| --- | --- | --- | --- |
| LOC | 3.6387 | - 6.2708 | 0.5022 |
| ORG | 1.2701 | - 0.5790 | 0.3049 |
| PER | 3.3895 | - 3.7817 | 0.2932 |
| O | 0.8832 | - 0.0988 | 0.0636 |

## 4. Analysis

Our diagnostic experiments, quantitatively grounded by the semantic drift metrics in Table 2, reveal the internal mechanics governing the model's ability to extend its capabilities. The analysis provides evidence for the independent processing of feature types and exposes the roles of representation overlap and background class plasticity.

### 4.1 The Independent Nature of Semantic and Pattern-based Features

Our first key observation comes from comparing the Baseline model (trained only on PER, LOC, ORG) with the Joint Training model (trained on all seven entity types). Contrary to the common expectation of performance degradation, the introduction of four new PII entities had no significant negative impact on the original, standard entity types.

This remarkable stability suggests that the features used by the model to identify the two groups of entities are largely independent. We hypothesize that the model learns to recognize the new PII entities (PHONE, EMAIL, IBAN, PDL) primarily through low-level lexical and morphological patterns, while the standard entities are identified using higher-level semantic and contextual cues. The Original vs. Joint data in Table 2 provides quantitative support for this hypothesis. The covariance drift for all old entities is low (0.22 for LOC and PER, 0.28 for ORG), a direct indication that learning the pattern-based PII entities did not disrupt the internal feature structures used to define the semantic entities. The concurrent negative variance change, especially for ORG (-4.78), further suggests that, in an ideal context, with a more diverse set of entities to distinguish against, the model learned more compact and well-defined representations for the old classes.

### 4.2. Representation Overlap: Diagnosing Vulnerabilities with Incremental Learning

While Joint Training reveals ideal coexistence, it does not expose the model's potential failure points. The Naive Incremental Learning experiment, which serves as the primary test in our ablation study, revealed critical vulnerability: While all old entities suffered from catastrophic forgetting, the damage was not uniform. As demonstrated in Table 1, in the naive incremental learning scenario, the LOC entity's performance collapsed disproportionately, dropping by over 20 F1 points, far exceeding the degradation of PER and ORG.

This disproportionate impact on the LOC entity indicates a representation overlap between it and the new PII entities. Unlike PER and ORG, which are almost purely semantic, LOC entities in our spoken-language data often contain structured, pattern-like elements such as street numbers or postal codes. These features resemble the morphological nature of phone numbers or IBANs. When the model updates its understanding of these patterns during incremental learning, the gradients cause collateral damage to the LOC class, which partially relies on these same underlying features.The Original vs. Naive comparison shows LOC's covariance drift surging to 1.00, indicating a significant alteration of its core representational structure.

Another notable observation across all naive learning comparisons (e.g., Original vs. Naive and Joint vs. Naive) is that PER consistently exhibits a larger variance expansion than LOC, yet its F1 score degradation is less severe. For instance, in Original vs. Naive, PER's variance expands by +6.84 compared to LOC's +4.59. This disparity can be explained by their differing failure modes. LOC's high covariance drift (1.00) suggests a structural corruption of its feature logic due to direct pattern overlap. In contrast, PER's lower covariance drift (0.46) suggests its core definition was better preserved, and its failure was primarily a case of boundary confusion, manifested as high representational uncertainty.

### 4.3 Reverse O-tag representation drift

Further analysis within our ablation study uncovered the pivotal and counter-intuitive role of the 'O' tag. The initial training phase creates a powerful bias, teaching the model to map PII-like patterns to the 'O' class. This creates a state of intransigence—confirmed by ablations where freezing all old classifier heads including 'O' completely stalled new entity learning. The drift metrics quantify this dynamic: in the Original vs. Naive experiment, the O-tag's representation undergoes a significant restructuring, evidenced by a high covariance drift (1.08) and a concurrent variance contraction (-1.20) as PII patterns are removed from its distribution.

The performance improvement observed in our Incremental with O-Tag Adaptation experiment allows us to diagnose this phenomenon. Allowing the 'O' tag classifier to remain trainable unlocks model plasticity and enables it to begin learning the new entity types. We interpret this as the resolution of a background class ambiguity. The model's learned associations between PII patterns and the 'O' class create classification conflicts when these patterns must be reassigned. Comparing the Original vs. Naive and Original vs. Freeze except O results, we see that allowing the O-tag to adapt while freezing other classifiers stabilizes the LOC representation (its covariance drift drops from 1.00 to 0.21), while the O-tag's own drift remains high (1.08). This demonstrates that updating the 'O' classifier allows the model to restructure its decision boundaries, selectively transferring patterns from background to entity classifications.

## 5. Conclusion

In this work, we investigated the dynamics of extending a pre-trained NER model with new, pattern-based PII entities in spoken-language context. Contrary to expectations of performance degradation, joint training incorporating new entities showed no significant negative impact on original semantic entities (PER, LOC, ORG). This stability stems from the independent nature of semantic and morphological features used to identify distinct entity groups.

Using an incremental learning framework as a diagnostic tool and the quantified semantic drift framework in Wu et al. (2025a), we conducted an ablation study that revealed two key insights into the model's internal mechanics. First, we identified a representation overlap affecting the LOC entity, whose hybrid semantic-morphological nature makes it uniquely vulnerable to interference from new pattern-based classes. Second, we demonstrated that the model's inability to learn new entities stemmed from a strong, pre-existing association of PII patterns with the 'O' tag. Enabling adaptation of the background class definition resolves this ambiguity and classification conflict while preserving existing knowledge.

Our findings contribute a deeper, more nuanced understanding of how transformer-based models handle evolving entity ontologies, shifting the focus from preventing catastrophic forgetting to understanding the conditions that enable learning in the first place.

### 6. Limitations and Future Work

While our study provides valuable insights, it presents several limitations that constrain the generalizability of findings. The analysis relies on a specific configuration of PII and standard entities within a proprietary dataset. While we believe the underlying principles of feature independence and background class adaptation are broadly applicable, validation across diverse datasets, languages, and entity configurations remains necessary. Additionally, our investigation focuses on adding pattern-based rather than semantic entities, potentially limiting insights for other expansion scenarios. The diagnostic approach prioritizes understanding over optimization, leaving room for developing improved incremental learning methods.

Our diagnostic findings suggest several promising research directions. The discovery of reverse O-tag drift reveals that incremental learning challenges extend beyond inter-entity interference to include complex dynamics between entity and background classifications. Future algorithms should model the bidirectional relationship where background class adaptation affects entity recognition and vice versa. Additionally, our finding that feature independence enables peaceful coexistence between entity types motivates broader investigation into the conditions under which different NER extensions succeed or fail. Systematic studies across various entity type combinations could reveal general principles for predicting and mitigating interference patterns. Finally, the diagnostic framework we employed—using incremental learning to expose model vulnerabilities—could be applied to other NER challenges, providing insights into model behavior that joint training alone cannot reveal.

## 7. References

Aleixo, Everton L., Juan G. Colonna, Marco Cristo, and Everlandio Fernandes. "Catastrophic forgetting in deep learning: A comprehensive taxonomy." *arXiv preprint arXiv:2312.10549*. 2023.

Aljundi, Rahaf, Francesca Babiloni, Mohamed Elhoseiny, Marcus Rohrbach, and Tinne Tuytelaars. "Memory aware synapses: Learning what (not) to forget." In *Proceedings of the European conference on computer vision (ECCV)*, pp. 139-154. 2018.

Bell, Samuel J., and Neil D. Lawrence. "Behavioral experiments for understanding catastrophic forgetting." *arXiv preprint arXiv:2110.10570*. 2021.

Chaudhry, Arslan, Marc'Aurelio Ranzato, Marcus Rohrbach, and Mohamed Elhoseiny. "Efficient lifelong learning with a-gem." *arXiv preprint arXiv:1812.00420*. 2018.

Davari, MohammadReza, Nader Asadi, Sudhir Mudur, Rahaf Aljundi, and Eugene Belilovsky. "Probing representation forgetting in supervised and unsupervised continual learning." In *Proceedings of the IEEE/CVF conference on computer vision and pattern recognition*, pp. 16712-16721. 2022.

Hou, Saihui, Xinyu Pan, Chen Change Loy, Zilei Wang, and Dahua Lin. "Learning a unified classifier incrementally via rebalancing." In *Proceedings of the IEEE/CVF conference on computer vision and pattern recognition*, pp. 831-839. 2019.

Kim, Joonkyu, Yejin Kim, and Jy-yong Sohn. "Measuring Representational Shifts in Continual Learning: A Linear Transformation Perspective." In *Forty-second International Conference on Machine Learning*. 2025.

Kirkpatrick, James, Razvan Pascanu, Neil Rabinowitz, Joel Veness, Guillaume Desjardins, Andrei A. Rusu, Kieran Milan et al. "Overcoming catastrophic forgetting in neural networks." In *Proceedings of the national academy of sciences* 114, no. 13: 3521-3526. 2017.

Lancheros, Brayan Stiven, Gloria Corpas Pastor, and Ruslan Mitkov. "Data augmentation and transfer learning for cross-lingual Named Entity Recognition in the biomedical domain." *Preprint at Research Square* [https://doi.org/10.21203/rs.3.rs-2557266/v1]. 2023.

Li, Zhizhong, and Derek Hoiem. "Learning without forgetting." In *IEEE transactions on pattern analysis and machine intelligence* 40, no. 12: 2935-2947. 2017.

Liu, Xialei, Marc Masana, Luis Herranz, Joost Van de Weijer, Antonio M. Lopez, and Andrew D. Bagdanov. "Rotate your networks: Better weight consolidation and less catastrophic forgetting." In *2018 24th international conference on pattern recognition (ICPR)*, pp. 2262-2268. IEEE, 2018.

Lopez-Paz, David, and Marc'Aurelio Ranzato. "Gradient episodic memory for continual learning." In *Advances in neural information processing systems* 30. 2017.

Ma, Chunwei, Zhanghexuan Ji, Ziyun Huang, Yan Shen, Mingchen Gao, and Jinhui Xu. "Progressive voronoi diagram subdivision enables accurate data-free class-incremental learning." In *The Eleventh International Conference on Learning Representations*. 2023a.

Ma, Ruotian, Xuanting Chen, Zhang Lin, Xin Zhou, Junzhe Wang, Tao Gui, Qi Zhang, Xiang Gao, and Yun Wen Chen. "Learning "O" helps for learning more: Handling the unlabeled entity problem for class-incremental NER." In *Proceedings of the 61st Annual Meeting of the Association for Computational Linguistics (Volume 1: Long Papers)*, pp. 5959-5979. 2023b.

Mallya, Arun, and Svetlana Lazebnik. "Packnet: Adding multiple tasks to a single network by iterative pruning." In *Proceedings of the IEEE conference on Computer Vision and Pattern Recognition*, pages 7765–7773. 2018.

McCloskey, Michael, and Neal J. Cohen. "Catastrophic interference in connectionist networks: The sequential learning problem." In *Psychology of learning and motivation*, vol. 24, pp. 109-165. Academic Press, 1989.

Monaikul, Natawut, Giuseppe Castellucci, Simone Filice, and Oleg Rokhlenko. "Continual learning for named entity recognition." In *Proceedings of the AAAI Conference on Artificial*

*Intelligence*, vol. 35, no. 15, pp. 13570-13577. 2021.

Pauline Béraud, Margaux Rioux, Michel Babany, Philippe de La Chevasnerie, Damien Theis, Giacomo Teodori, Chloé Pinguet, Romane Rigaud and François Leclerc. "An Explainable Machine Learning Approach for Energy Forecasting at the Household Level" in arXiv preprint arXiv:2410.14416v1

Rebuffi, Sylvestre-Alvise, Alexander Kolesnikov, Georg Sperl, and Christoph H. Lampert. "icarl: Incremental classifier and representation learning." In *Proceedings of the IEEE conference on Computer Vision and Pattern Recognition*, pp. 2001-2010. 2017.

Shin, Hanul, Jung Kwon Lee, Jaehong Kim, and Jiwon Kim. "Continual learning with deep generative replay." In *Advances in neural information processing systems* 30. 2017.

Robins, Anthony. "Catastrophic forgetting, rehearsal and pseudorehearsal." In *Connection Science* 7, no. 2: 123-146. 1995.

Rusu, Andrei A., Neil C. Rabinowitz, Guillaume Desjardins, Hubert Soyer, James Kirkpatrick, Koray Kavukcuoglu, Razvan Pascanu, and Raia Hadsell. "Progressive neural networks." *arXiv preprint arXiv:1606.04671*. 2016.

Tian, Songsong, Lusi Li, Weijun Li, Hang Ran, Xin Ning, and Prayag Tiwari. "A survey on few-shot class-incremental learning." In *Neural Networks* 169: 307-324. 2024.

Serrà, Joan, Didac Suris, Marius Miron, and Alexandros Karatzoglou. "Overcoming catastrophic forgetting with hard attention to the task." In *International conference on machine learning*, pp. 4548-4557. PMLR, 2018.

Wang, Fu-Yun, Da-Wei Zhou, Han-Jia Ye, and De-Chuan Zhan. "Foster: Feature boosting and compression for class-incremental learning." In *European conference on computer vision*, pp. 398-414. Cham: Springer Nature Switzerland, 2022a.

Wang, Liyuan, Xingxing Zhang, Hang Su, and Jun Zhu. "A comprehensive survey of continual learning: Theory, method and application." In *IEEE transactions on pattern analysis and machine intelligence* 46, no. 8: 5362-5383. 2024

Wang, Rui, Tong Yu, Handong Zhao, Sungchul Kim, Subrata Mitra, Ruiyi Zhang, and Ricardo Henao. "Few-shot class-incremental learning for named entity recognition." In *Proceedings of the 60th Annual Meeting of the Association for Computational Linguistics* (Volume 1: Long Papers), pp. 571-582. 2022b.

Wang, Zhen, Liu Liu, Yiqun Duan, Yajing Kong, and Dacheng Tao. "Continual learning with lifelong vision transformer." In *Proceedings of the IEEE/CVF Conference on Computer Vision and Pattern Recognition*, pp. 171-181. 2022c.

Wang, Zifeng, Zizhao Zhang, Chen-Yu Lee, Han Zhang, Ruoxi Sun, Xiaoqi Ren, Guolong Su, Vincent Perot, Jennifer Dy, and Tomas Pfister. "Learning to prompt for continual learning." In *Proceedings of the IEEE/CVF conference on computer vision and pattern recognition*, pp. 139-149. 2022d.

Wiewel, Felix, and Bin Yang. "Localizing catastrophic forgetting in neural networks." *arXiv preprint arXiv:1906.02568*. 2019.

Wu, Fangwen, Lechao Cheng, Shengeng Tang, Xiaofeng Zhu, Chaowei Fang, Dingwen Zhang, and Meng Wang. "Navigating Semantic Drift in Task-Agnostic Class-Incremental Learning." *arXiv preprint arXiv:2502.07560*. 2025a.

Wu, Zihao, YongXiang Hua, Yongxin Zhu, Fang Zhang, and Linli Xu. "Dynamic Prefix as Instructor for Incremental Named Entity Recognition: A Unified Seq2Seq Generation Framework." In *Findings of the Association for Computational Linguistics: ACL 2025*, pp. 3294-3306. 2025b.

Xia, Yu, Quan Wang, Yajuan Lyu, Yong Zhu, Wenhao Wu, Sujian Li, and Dai Dai. "Learn and review: Enhancing continual named entity recognition via reviewing synthetic samples." In *Findings of the association for computational linguistics*: ACL 2022, pp. 2291-2300. 2022.

Yan, Shipeng, Jiangwei Xie, and Xuming He. "Der: Dynamically expandable representation for class incremental learning." In *Proceedings of the IEEE/CVF conference on computer vision and pattern recognition*, pp. 3014-3023. 2021.

Zenke, Friedemann, Ben Poole, and Surya Ganguli. "Continual learning through synaptic intelligence." In *International conference on machine learning*, pp. 3987-3995. PMLR, 2017.

Zhao, Li-Jun, Zhen-Duo Chen, Yongxin Wang, Xin Luo, and Xin-Shun Xu. "Attraction Diminishing and Distributing for Few-Shot Class-Incremental Learning." In *Proceedings of the Computer Vision and Pattern Recognition Conference*, pp. 25657-25666. 2025.

Zheng, Junhao, Shengjie Qiu, and Qianli Ma. "Learn or recall? revisiting incremental learning with pre-trained language models." *arXiv preprint arXiv:2312.07887*. 2023.

Zhou, Da-Wei, Qi-Wei Wang, Zhi-Hong Qi, Han-Jia Ye, De-Chuan Zhan, and Ziwei Liu. "Class-incremental learning: A survey." In *IEEE Transactions on Pattern Analysis and Machine Intelligence*. 2024.

## 8. Data

Data used was generated at papernest (French scale up helping people moving to subscribe, cancel and manage their household subscriptions (electricity, gas, home insurance, broadband and mobile contracts), located at 157 boulevard Macdonald 75019 Paris).

Further information can be found at https://www.papernest.com, https://www.papernest.es & https://www.papernest.it